%% file: main.tex
\documentclass[conference]{IEEEtran}
\IEEEoverridecommandlockouts
\usepackage{cite}
\usepackage{amsmath,amssymb,amsfonts}
\usepackage{algorithmic}
\usepackage{graphicx}
\usepackage{textcomp}
\usepackage{xcolor}

\usepackage{booktabs}
\usepackage{comment}

\def\BibTeX{{\rm B\kern-.05em{\sc i\kern-.025em b}\kern-.08em
    T\kern-.1667em\lower.7ex\hbox{E}\kern-.125emX}}
\begin{document}

\title{Adaptable and Reliable Text Classification using Large Language Models\\

}

\author{\IEEEauthorblockN{Zhiqiang Wang, Yiran Pang, Yanbin Lin, Xingquan Zhu}
\IEEEauthorblockA{\textit{Department of Electrical Engineering and Computer Science, Florida Atlantic University}\\
Boca Raton, FL 33431, USA \\
{\{zwang2022, ypang2022,  liny2020, xzhu3\}@fau.edu}
}}

\maketitle

\begin{abstract}
Text classification is fundamental in Natural Language Processing (NLP), and the advent of Large Language Models (LLMs) has revolutionized the field. This paper introduces an adaptable and reliable text classification paradigm, which leverages LLMs as the core component to address text classification tasks. Our system simplifies the traditional text classification workflows, reducing the need for extensive preprocessing and domain-specific expertise to deliver adaptable and reliable text classification results. We evaluated the performance of several LLMs, machine learning algorithms, and neural network-based architectures on four diverse datasets. Results demonstrate that certain LLMs surpass traditional methods in sentiment analysis, spam SMS detection, and multi-label classification. Furthermore, it is shown that the system's performance can be further enhanced through few-shot or fine-tuning strategies, making the fine-tuned model the top performer across all datasets. Source code and datasets are available in this GitHub repository: https://github.com/yeyimilk/llm-zero-shot-classifiers.
\end{abstract}


\input{sec/introduction} 
\input{sec/background}

\input{sec/methodology}

\input{sec/dataset}
\input{sec/experiments}
\input{sec/discussion}

\section{Conclusion and Future Work}

In conclusion, our study has demonstrated the potential of LLMs as effective text classifiers, often surpassing traditional ML and NN approaches. Strategic fine-tuning has proven to be an influential method for enhancing LLMs' domain-specific performance.

Our findings highlight the adaptability of LLMs in streamlining the text classification process by eliminating the need for extensive data preprocessing. This adaptability is particularly beneficial for small businesses looking for cost-effective solutions to integrate intelligent text classification without the requirement for deep ML or DL expertise. By democratizing access to advanced AI technology, LLMs empower organizations with limited resources to leverage sophisticated NLP tools. Businesses can efficiently process user feedback, enhance spam detection mechanisms, and automate workflows with minimal engineering effort, demonstrating the reliable and high-performance standards of our approach.

For future work, we aim to focus on making the system more reliable. Directions include but are not limited to employing a secondary LLM to process initial classification results, which could reduce U/E rates. 

\section*{Acknowledgements}
This study is supported by the U.S. National Science Foundation under grant Nos. IIS-2236579, IIS-2302786 and IOS-2430224.
\bibliographystyle{IEEEtran}
\bibliography{ref}

\end{document}

%% file: sec/introduction.tex
\section{Introduction}

Text classification is a core task in natural language processing (NLP), with applications ranging from sentiment analysis to question answering \cite{liu2022sentiment, chen2020dirichlet, minaee2021deep}. Traditional machine learning (ML) methods, such as logistic regression and Naive Bayes \cite{sarker2021machine, wang2019survey}, have been widely employed. However, these approaches often require extensive labeled datasets and are limited in adapting to unseen data or emerging categories, thus posing challenges in dynamic real-world environments.

\begin{figure}[h]
  \centering
  \includegraphics[width=0.45\textwidth]{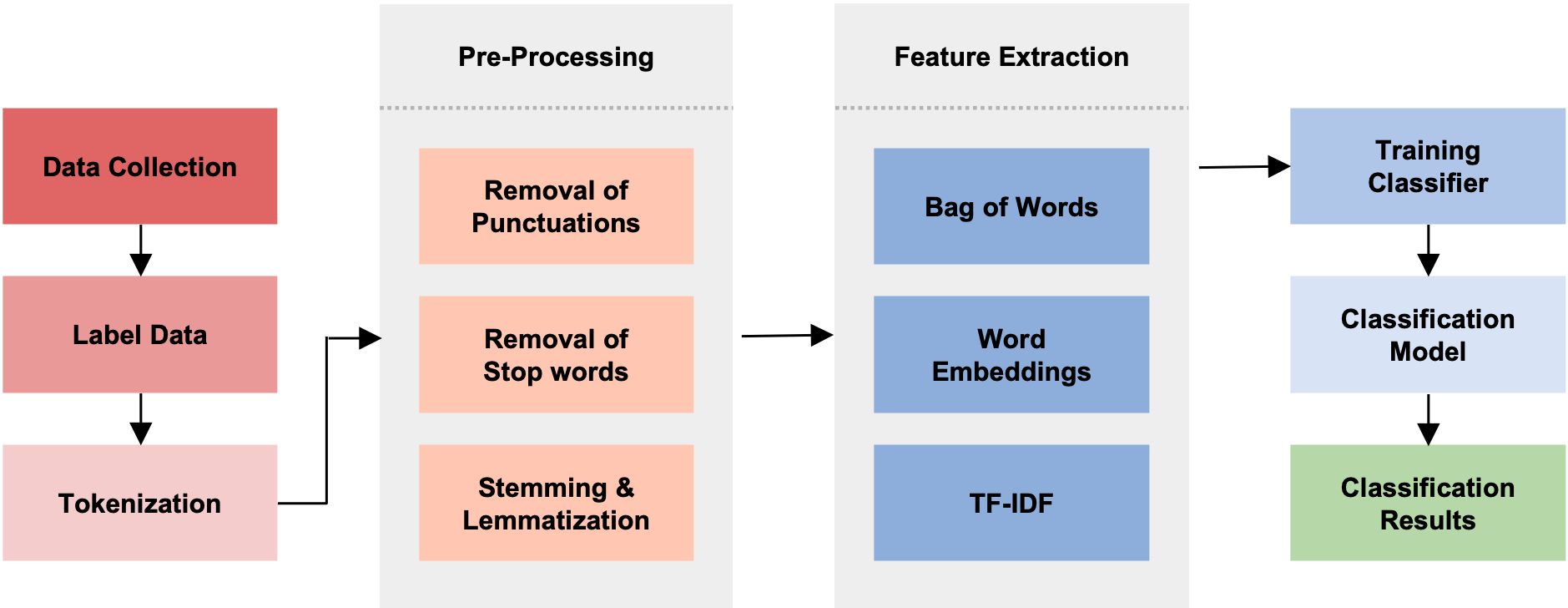}
    \caption{Traditional text classification flow}
    \label{fig:traditional_flow}
\end{figure}

\begin{figure}[h]
  \centering
  \includegraphics[width=0.45\textwidth]{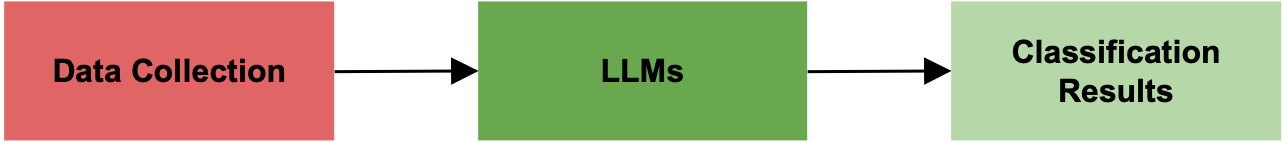}
    \caption{LLMs' zero-shot text classification simple flow}
    \label{fig:llms_flow}
\end{figure}

The emergence of large language models (LLMs) based on Transformer architectures, such as PaLM \cite{chowdhery2023palm}, LLaMA \cite{touvron2023llama}, and GPT \cite{radford2018improving}, has transformed the landscape of text classification. Unlike traditional approaches, as shown in Figure \ref{fig:traditional_flow}, which require complex, multi-step pipelines for data preprocessing and feature extraction, LLMs leverage their extensive pre-training to handle these tasks internally. In contrast to the more labor-intensive traditional method, this shift reduces the need for manual intervention and allows the models to generalize more effectively across various domains. As illustrated in Figure \ref{fig:llms_flow}, the LLM-based approach condenses the workflow into three main stages: data collection, feeding data directly into the LLM, and receiving classification outputs.

\begin{figure*}[th]
  \centering
  \includegraphics[width=1\textwidth]{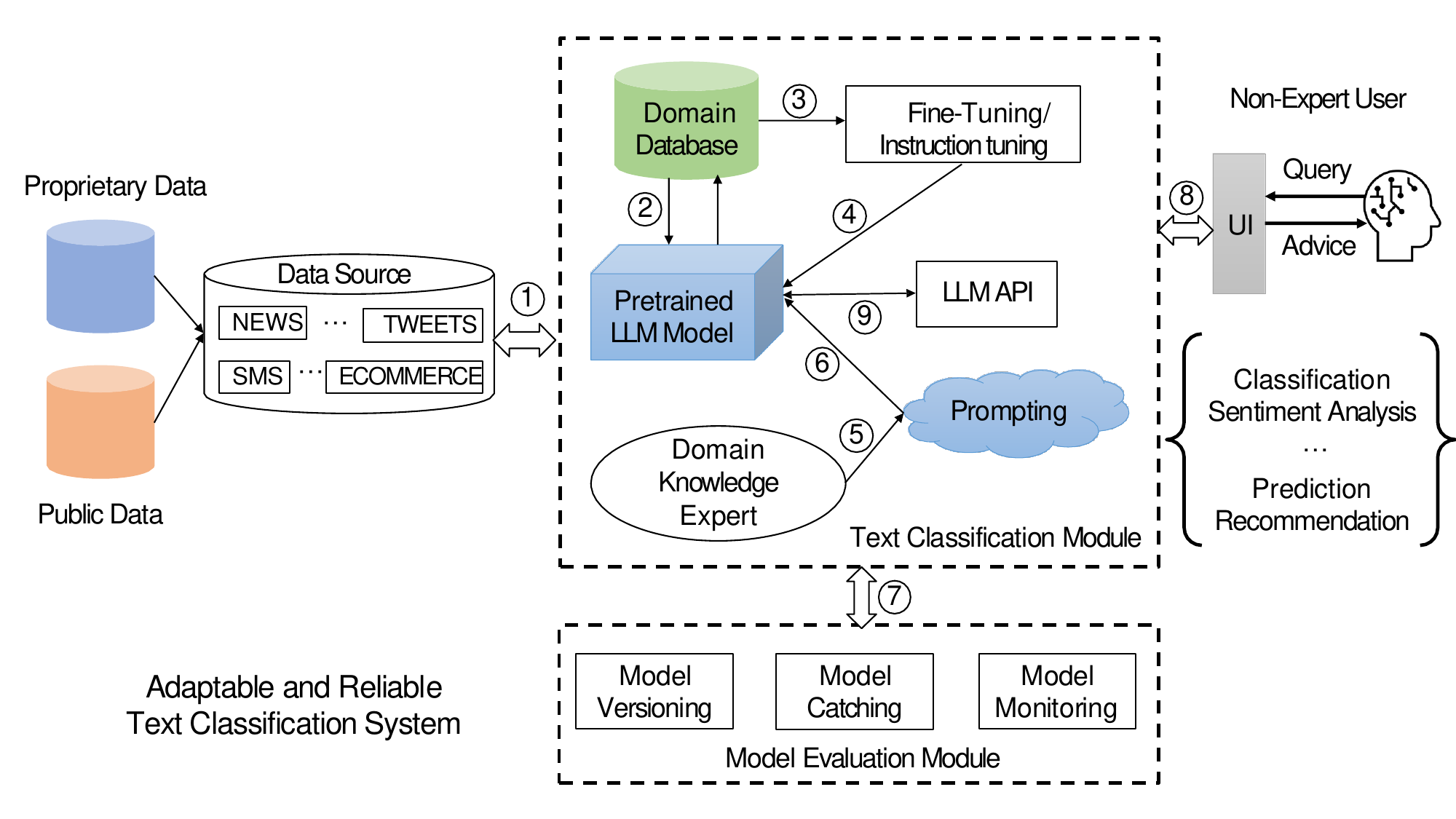}
    \caption{Framework of our adaptable and reliable text classification system. The steps of the framework can be included as (1) collect data from the data source to establish the domain database; (2) send domain-specific data to the pre-trained LLM model, like GPT-4, Llama-3 and so on; (3) using a few domain-specific data to do fine-tuning or instruction tuning (4) apply the fine-tuning or instruction tuning to the pre-trained LLM model; (5) (optional) utilize domain knowledge to set up the prompts to elevate LLM performance; (6) apply prompts in the pre-trained model; (7) evaluate the whole system's performance; (8) non-expert users query tasks through user interface to the system; (Tasks may include classification, sentiment analysis, prediction, recommendation and so on. In this paper, we take the multi-class classification and sentiment analysis as examples.) (9) LLM API interacts with User interface and the pre-trained LLM model, advising on the user interface.}
    \label{fig: expert_system}
\end{figure*} 

Despite these advancements, deploying LLMs in real-world text classification tasks still presents particular challenges. For instance, LLMs must maintain high reliability across diverse and unpredictable environments, ensuring robustness even when dealing with domain shifts or rare categories \cite{bețianu2024dallmi, yan2023autocast}. Additionally, the responsibility of these models is increasingly essential, as fairness, transparency, and ethical considerations come to the forefront when implementing LLMs in decision-making systems~\cite{Wu2024llm}.

In response to these challenges, we propose a novel text classification framework that harnesses the strengths of LLMs while addressing critical aspects of adaptability. Our framework, illustrated in Figure \ref{fig: expert_system}, integrates LLMs at the core of the classification workflow, significantly simplifying the process. It allows non-expert users to access high-performing classification systems with minimal effort, reducing the extensive preprocessing and feature engineering traditionally required.

Our main contributions are as follows:

\begin{itemize}
    \item This paper proposes a text classification system using LLMs to replace traditional text classifiers. This system simplifies the conventional text classification process, lowering technical barriers and eliminating the need for domain experts to perform complex preprocessing and algorithm design. This approach is crucial for rapid deployment and scalable applications, especially for small businesses needing deep ML or DL expertise.
    \item We introduce a new performance evaluation metric, the Uncertainty/Error Rate (U/E rate). This metric supplements traditional accuracy and F1 scores, providing a more comprehensive evaluation of a model's performance under unknown or uncertain conditions and emphasizing the LLMs' reliability in real-world applications.
    \item We compare the performance of LLMs with traditional ML and NN models across multiple datasets. After learning from a few samples or fine-tuning, the results show that LLMs outperform in various text classification tasks. This finding confirms the versatility and efficiency of LLMs.
\end{itemize}

%% file: sec/background.tex
\section{Background and Related Work}

\subsection{Traditional Text Classification Approaches}
Text classification has evolved through various machine learning (ML) methods, each with its strengths and limitations. Early rule-based approaches, such as decision trees like C4.5 \cite{quinlan2014c4}, were simple but prone to overfitting and lacked flexibility. Probability-based models, such as Multinomial Naive Bayes (MNB) \cite{xu2018bayesian} and Hidden Markov Models \cite{rabiner1989tutorial}, improved generalization, particularly in tasks like spam detection and speech recognition. Geometry-based methods, including support vector machine (SVM) \cite{joachims2002learning}, handled high-dimensional data but struggled with large datasets. Finally, statistical methods like K-nearest neighbors (KNN) \cite{guo2003knn} and Logistic Regression (LG) \cite{genkin2007large} provided effective solutions but required extensive preprocessing and often faltered with nonlinear data.

\subsection{Deep Learning Approaches}
Deep learning (DL) has become a key technology in text classification, capable of handling complex language features. Convolutional neural networks (CNNs) text classification models \cite{kim2014convolutional} capture local textual features through convolutional layers. LSTM \cite{sutskever2014sequence} and GRU \cite{chung2014empirical}, as optimized versions of RNNs, are particularly effective in addressing long-distance dependencies in text. \cite{zhou2024clcrnet} proposes an optical character recognition and classification method for cigarette laser code recognition, using a convolutional recurrent neural network to extract image features and utilizing BiLSTM for text classification. Transformer models, like BERT \cite{devlin2018bert}, achieve remarkable results in various NLP tasks by utilizing self-attention mechanisms. Specifically, the BERT model demonstrates powerful capabilities in text classification tasks. \cite{raza2024nbias} proposes the text recognition framework Nbias for detecting and eliminating biases, including data, corpus construction, model development, and evaluation layers. The dataset is collected from various fields, and a transformer-based token classification model is applied. \cite{zhou2024cdgan} proposes a semi-supervised generative adversarial learning method that improves the model's classification performance with limited annotated data through generative adversarial networks. \cite{jamshidi2024effective} introduces a hybrid model that combines BERT, LSTM, and Decision Templates (DT) for IMDB and Drug Review classification. However, these methods typically require substantial data for training, often necessitating extensive datasets to achieve optimal performance. This reliance on large training sets can pose challenges, especially when collecting or labeling data is difficult or impractical.

\subsection{LLM Approaches}
LLMs represent a significant advancement in the field of text classification, building on the DL foundations that have revolutionized NLP. These models, which include notable examples such as GPT \cite{brown2020language}, T5 \cite{raffel2020exploring}, RWKV \cite{peng2023rwkv}, Mamba \cite{gu2023mamba}, Gemini \cite{team2023gemini}, PaLM \cite{chowdhery2023palm}, Llama \cite{touvron2023llama}, and Claude \cite{anthropic2024claude}, leverage massive amounts of data and extensive training regimes to understand and generate human-like text. Their ability to capture nuanced language patterns and context makes them highly effective for text classification tasks across various domains.

Recent studies have begun to explore the practical applications of LLMs in specialized fields. For instance, \cite{chae2023large} study the application of LLMs in sociological text classification, demonstrating their potential in social science research. \cite{loukas2023making} examine the performance and cost trade-offs when employing LLMs for text classification, focusing on financial intent detection datasets. Another study by \cite{10386911} investigates the effect of fine-tuning LLMs on text classification tasks within legal document review, highlighting how domain-specific adjustments can enhance model performance. Furthermore, the research identified as \cite{10386772} refines LLM performance on multi-class imbalanced text classification tasks through oversampling techniques, addressing one of the common challenges in ML. 

Despite their strengths, there remains a gap in making LLMs accessible to users without deep technical expertise. Our system addresses this by leveraging pre-trained LLMs as out-of-the-box classifiers that require minimal adaptation. This system democratizes access to advanced NLP tools, offering scalable solutions for diverse applications without the steep learning curve typically associated with LLM deployment.

%% file: sec/methodology.tex
\section{Methodology}

\subsection{Adaptable and Reliable System}

Our proposed system integrates LLMs to refine the traditional text classification system, as illustrated in Figure \ref{fig: expert_system} based on our previous work \cite{wang2023large}. The framework of our system presents a comprehensive strategy that capitalizes on the strengths of LLMs while mitigating their traditional limitations.

Initially, our system aggregates data from many sources, either public or private. Unlike traditional ML/NN methods, which often require extensive retraining or fine-tuning when confronted with new data types, our LLM-based system can effectively adapt to these varied inputs without additional training. This versatility is one of the key strengths of our approach.

Subsequently, the system harnesses domain-specific data through zero-shot prompting or few-shot learning techniques or by fine-tuning a pre-trained LLM. This adaptive phase meticulously tailors the LLM's capabilities to suit the target domain's particular linguistic features and contextual subtleties, thereby bolstering accuracy and relevance for classification tasks.

Furthermore, the involvement of domain knowledge is crucial but optional. They configure the system by establishing customized prompts that direct the LLM toward generating pertinent and contextually aware responses. This human-in-the-loop methodology guarantees that the system adheres to specific domain requirements and can adeptly manage intricate query scenarios.

Additionally, an LLM API serves as an intermediary between the model and user interface, enabling seamless real-time interactions. Through this user-friendly interface, users without expertise can effortlessly query the system for advice, classification results, sentiment assessments, predictions, or recommendations based on their input.

Lastly, our system incorporates an evaluation subsystem dedicated to continuously monitoring LLM performance. It scrutinizes accuracy and error rates while observing model behavior over time. Such vigilance facilitates perpetual enhancements and updates via model versioning and caching strategies.

By amalgamating these components, our system simplifies and elevates text classification processes in terms of adaptability and precision. It significantly diminishes reliance on domain knowledge for complex preprocessing or algorithmic design tasks—thereby democratizing access to cutting-edge NLP technologies across various sectors such as e-commerce and social media analytics.

\subsection{Evaluation metrics}

We utilized several key metrics to assess the performance of LLMs as text classifiers. These metrics provide insights into the accuracy, precision, recall, and stability of the LLMs in handling classification tasks.

\subsubsection{Accuracy}

This metric measures the proportion of correct predictions made by the model out of all predictions. It is calculated using the formula:

\begin{equation}
\text{ACC} = \frac{TP + TN}{TP + TN + FP + FN}
\label{equ: acc}
\end{equation}

where $TP$ is the number of true positives, $TN$ is the number of true negatives, $FP$ is the number of false positives, $FN$ is the number of false negatives.

\subsubsection{F1 Score}

The F1 score is a harmonic mean of precision and recall, providing a balance between them. It is particularly useful when dealing with imbalanced classes. 





\begin{equation}
F1 = \frac{\text{2TP}}{\text{2TP} + \text{FP} + \text{FN}}
\end{equation}

\subsubsection{U/E Rate}

We propose a novel metric called Uncertainty/Error Rate (U/E rate) to evaluate the stability and reliability of LLM outputs. This metric quantifies the frequency at which an LLM either refuses to classify content or provides an output deemed unrelated or beyond its capabilities. The U/E rate is defined as:

\begin{equation}
\text{U/E} = \frac{U + E}{N}
\label{equ: ue}
\end{equation}

where $U$ is the number of uncertain outputs (e.g., refusals to classify), $E$ is the number of erroneous outputs (e.g., unrelated or hallucinated results), and $N$ is the total number of test samples. 

The U/E rate complements traditional performance metrics by highlighting instances where LLMs exhibit behavior divergent from deterministic ML/NN models, such as refusing to analyze content or producing hallucinated results.

By employing these evaluation metrics, we aim to provide a multifaceted view of LLM performance that encompasses traditional aspects like accuracy and F1 score and novel considerations introduced by their unique operational characteristics.

%% file: sec/dataset.tex
\section{Dataset}

Four datasets include varying lengths of text inputs (from short tweets to longer reviews), domain-specific language usage (as seen in economic texts), diverse sentiment expressions (ranging from public health concerns to consumer products), and practical applications such as spam filtering, were employed to evaluate the LLMs' adaptability and reliability in handling text classification tasks.

\subsection{COVID-19-related Tweets Dataset}

\begin{table}[h]
\caption{\small{COVID-19-related Tweets Dataset Statistics}}
\centering
\label{tab:ds_tweets}
\renewcommand{\arraystretch}{1.4}
\begin{tabular}{lcccc}
\hline
    & Negative & Neutral & Positive & Total \\ \hline
    Train  & 15398 & 7712 & 18046 & 41156 \\
    Test   & 1633 & 619 & 1546 & 3798 \\
    \hline
\end{tabular}
\end{table}

The first dataset consists of tweets related to the COVID-19 pandemic, curated by \cite{gabrielpreda_2020}. As shown in Table \ref{tab:ds_tweets}, it comprises a total of 41,156 training instances and 3,798 test instances, categorized into negative, neutral, and positive sentiments.

\subsection{Economic Texts Dataset}

\begin{table}[h]
\caption{\small{Economic texts Dataset Statistics}}
\centering
\label{tab:ds_economic}
\renewcommand{\arraystretch}{1.4}
\begin{tabular}{lcccc}
\hline
    & Negative & Neutral & 2 Positive & Total \\ \hline
    Train  & 483 & 2302 & 1091 & 3876 \\
    Test   & 121 & 576 & 272  & 969 \\
    \hline
\end{tabular}
\end{table}

The second dataset includes economic texts compiled by \cite{malo2014good}, designed for sentiment analysis within the financial domain. The dataset contains 3,876 training samples and 969 test samples distributed across negative, neutral, and positive classes, as detailed in Table \ref{tab:ds_economic}. This dataset includes 5 levels of sentiment, which were merged into 3 levels in this study.

\subsection{E-commerce Texts Dataset}

\begin{table}[h]
\caption{\small{E-commerce texts Dataset Statistics}}
\centering
\label{tab:ds_ecommerce}
\renewcommand{\arraystretch}{1.4}
\begin{tabular}{l@{\hspace{2mm}}c@{\hspace{2mm}}c@{\hspace{2mm}}c@{\hspace{2mm}}c@{\hspace{2mm}}c}
\hline
    & Household &  Books & C\&A & Electronics & Total \\ \hline
    Train  & 15449 & 9456 & 6936 & 8497 & 40338 \\
    Test   & 3863 & 2364 & 1734 & 2124 & 10085 \\
    \hline
\end{tabular}
\end{table}

For multi-class classification tasks beyond binary or ternary sentiment analysis, we utilize an e-commerce texts dataset provided by \cite{gautam_2019_335582}. The training set includes 40,338 instances, while the test set contains 10,085 instances, as outlined in Table \ref{tab:ds_ecommerce}.

\subsection{SMS Spam Collection Dataset}

\begin{table}[h]
\caption{\small{SMS Spam collection Statistics}}
\centering
\label{tab:ds_sms}
\renewcommand{\arraystretch}{1.4}
\begin{tabular}{lcccc}
\hline
    & Normal & Spam & Total \\ \hline
    Train  & 3859 & 598 & 4457 \\
    Test   & 966 & 149 & 1115 \\
    \hline
\end{tabular}
\end{table}

Lastly, we incorporate an SMS Spam Collection dataset assembled by \cite{misc_sms_spam_collection_228} to evaluate spam detection performance. This binary classification task involves distinguishing between normal messages and spam with a total of 4,457 training messages and 1,115 test messages presented in Table \ref{tab:ds_sms}.

%% file: sec/experiments.tex
\section{Experimental Results}

\subsection{Experiment Setup}
Our experiment setup is designed to evaluate the performance of various models across different categories, ensuring a comprehensive analysis of the proposed methods. The models are categorized as follows:

\begin{itemize}
    \item \textbf{Traditional ML Algorithms:} This category includes MNB, LG, RF, DT, and KNN.
    \item \textbf{NN Architectures:} We utilize advanced deep neural network models such as RNN, LSTM, and GRU.
    \item \textbf{Zero-shot Learning (ZSL) Models:} We explore zero-shot learning capabilities using transformer-based models, including BART (facebook/bart-large-mnli) and DeBERTa (microsoft/deberta-large-mnli).
    \item \textbf{LLMs:} State-of-the-art LLMs including closed source models: \textbf{GPT-3.5}(gpt-3.5-turbo-0125), \textbf{GPT-4 }(gpt-4-1106-preview), \textbf{Gemini-pro}, and open source models: \textbf{Llama3-8B}(Llama3-8B-Instruct), Qwen-Chat(7B and 14B), and Vicuna-v1.5(7B and 13B) were assessed.
\end{itemize}

To maintain consistency in evaluation, the input processing was standardized for all traditional ML algorithms and NN architectures. Each model receives the same processed text derived from a uniform raw text processing pipeline applied to training and testing datasets. This standardization ensures that any observed variations in performance can be attributed more directly to the intrinsic capabilities of each model rather than disparities in input processing.

Conversely, unprocessed raw text from the testing dataset was used to fully leverage their natural language understanding abilities for zero-shot learning models and LLMs. It is important to note that this testing dataset remains consistent across all model types to provide a fair comparison.

In addition to these measures, we implemented a sampling strategy for dataset selection that respects the original label distribution within both training and test sets:

\begin{itemize}
    \item  For datasets with more than 10,000 instances in their training set, only 10,000 instances were selected while preserving the original label distribution proportionally through stratified sampling.
 
    \item  Similarly, for test sets with more than 800 instances, only 800 instances were chosen based on their original label distribution.

\end{itemize}

This approach ensures that smaller datasets are fully represented while larger ones are sampled appropriately without introducing bias or altering their inherent class distributions.

Furthermore, when configuring prompts for LLMs within the experiments, uniformity is ensured by keeping prompts identical across different LLMs for the same dataset. When dealing with different datasets, a consistent core structure is maintained within prompts—only adjusting labels and dataset names as necessary—to minimize variability due to prompt differences.


\subsection{Experimental Results}

Table \ref{tab: twitters}, \ref{tab:e_commerce}, \ref{tab: economic_texts} and \ref{tab: sms_spam} present the experimental results for all the models. Notably, when employing few-shot strategies or fine-tuning, they are indicated by ``(S)" and ``(F)," respectively.

\begin{table}[h]
\caption{\small{COVID-19-RELATED TWEETS Sentiment classification results}}
\centering
\label{tab: twitters}
\renewcommand{\arraystretch}{1.4}
\begin{tabular}{l@{\hspace{2mm}}c@{\hspace{2mm}}c@{\hspace{2mm}}c@{\hspace{0mm}}}
\hline
Model & ACC($\uparrow$) & F1($\uparrow$) & U/E($\downarrow$) \\ 
\hline
MNB  & 0.4037 & 0.3827 & - \\
LR   & 0.3875 & 0.3131 & - \\
RF   & 0.4462 & 0.3633 & - \\
DT   & 0.4037 & 0.3416 & - \\
KNN  & 0.3825 & 0.3481 & - \\
\hline
GRU  & 0.6913 & 0.6324 & - \\
LSTM & 0.6687 & 0.6312 & - \\
RNN  & 0.6600 & 0.6332 & - \\
\hline
BART     & 0.5138 & 0.3638 & - \\
DeBERTa  & 0.5375 & 0.3804 & - \\
\hline
GPT-3.5                 & 0.5550 & 0.5435 & 0.0000 \\
GPT-4                   & 0.5100 & 0.5054 & 0.0000 \\
Gemini-pro              & 0.5025 & 0.5105 & 0.0388 \\
Llama-3-8B              & 0.5112 & 0.5149 & 0.0013 \\
Qwen-7B                 & 0.4913 & 0.4689 & 0.0025 \\
Qwen-14B                & 0.4562 & 0.4569 & 0.0100 \\
Vicuna-7B               & 0.3600 & 0.3403 & 0.0000 \\
Vicuna-13B              & 0.5050 & 0.4951 & 0.0013 \\
\hline
Gemini-pro(S)      & 0.4888\colorbox{red!25}{(-0.014)}     & 0.4880\colorbox{red!25}{(-0.022)}     & 0.0375\colorbox{green!25}{(-0.001)}   \\
Llama-3-8B(S)      & 0.5363\colorbox{green!25}{(+0.025)}   & 0.5298\colorbox{green!25}{(+0.015)}   & 0.0000\colorbox{green!25}{(-0.001)}   \\
Qwen-7B(S)         & 0.3900\colorbox{red!50}{(-0.101)}     & 0.3519\colorbox{red!50}{(-0.117)}     & 0.0150\colorbox{red!25}{(+0.012)}     \\
Qwen-14B(S)        & 0.4575\colorbox{green!25}{(+0.001)}   & 0.4556\colorbox{red!25}{(-0.001)}     & 0.0037\colorbox{green!25}{(-0.006)}   \\
Vicuna-7B(S)       & 0.3700\colorbox{green!25}{(+0.010)}   & 0.3362\colorbox{red!25}{(-0.004)}     & 0.0013\colorbox{red!25}{(+0.001)}     \\
Vicuna-13B(S)      & 0.5050\colorbox{green!0}{(+0.000)}   & 0.4951\colorbox{green!0}{(+0.000)}   & 0.0000\colorbox{green!25}{(-0.001)}   \\
\hline
Llama-3-8B(F) & 0.4675\colorbox{red!25}{(-0.044)} & 0.4910\colorbox{red!25}{(-0.024)} & 0.1175\colorbox{red!50}{(+0.116)} \\

Qwen-7B(F)  & \textbf{0.8388}\colorbox{green!75}{(+0.348)} & \textbf{0.8433}\colorbox{green!75}{(+0.374)} & 0.0000\colorbox{red!0}{(+0.000)} \\
\bottomrule
\multicolumn{4}{l}{\footnotesize S: with few shot strategy; F: with fine-tuned strategy} \\

\end{tabular}
\end{table}

Table \ref{tab: twitters} presents results for the COVID-19-related text dataset. All models have relatively low performance except the fine-tuned LLM model of Qwen-7B, which performed the best in all metrics with $0.8388$ in accuracy and $0.8433$ F1 score and clearly provided all the answers.

Traditional algorithms show poor accuracy and F1 scores. In contrast, NN-based models demonstrate superior performance, with GRU leading in both ACC and F1 metrics. Among LLMs, before fine-tuning, GPT-3.5 exhibits the highest ACC and F1 scores, outperforming other LLMs, including GPT-4, while the performance is below NN methods. However, once the fine-tuning method was employed, the Qwen-7B(F) outperformed all the other models, including GRU's best model. The best accuracy increased from 0.6913, performed by GRU, to 0.8388, and the F1 score from 0.63332, performed by RNN, to 0.8433. 

\begin{table}[h]
\caption{\small{E-Commercial Product text classification results}}
\centering
\label{tab:e_commerce}
\renewcommand{\arraystretch}{1.4}
\begin{tabular}{l@{\hspace{2mm}}c@{\hspace{2mm}}c@{\hspace{2mm}}c@{\hspace{0mm}}}
\hline
Model & ACC($\uparrow$) & F1($\uparrow$) & U/E($\downarrow$) \\ 
\hline
MNB  & 0.2562 & 0.2384 & - \\
LR   & 0.3825 & 0.2873 & - \\
RF   & 0.4875 & 0.3958 & - \\
DT   & 0.4263 & 0.4165 & - \\
KNN  & 0.3762 & 0.3414 & - \\
\hline
GRU  & 0.9387 & 0.9383 & - \\
LSTM & 0.9363 & 0.9398 & - \\
RNN  & 0.8975 & 0.9010 & - \\
\hline
BART     & 0.7175 & 0.7246 & - \\
DeBERTa  & 0.6025 & 0.6121 & - \\
\hline
GPT-3.5                 & 0.9125 & 0.9152 & 0.0063 \\
GPT-4                   & 0.9137 & 0.9221 & 0.0088 \\
Gemini-pro              & 0.8775 & 0.8873 & 0.0100 \\
Llama-3-8B              & 0.9113 & 0.9112 & 0.0000 \\
Qwen-7B                 & 0.5850 & 0.6584 & 0.1850 \\
Qwen-14B                & 0.6575 & 0.6843 & 0.0800 \\
Vicuna-7B               & 0.7100 & 0.7164 & 0.0050 \\
Vicuna-13B              & 0.8363 & 0.8503 & 0.0138 \\
\hline
Gemini-pro(S)      & 0.8862\colorbox{green!25}{(+0.009)}   & 0.8963\colorbox{green!25}{(+0.009)}   & 0.0100\colorbox{red!0}{(+0.000)}     \\
Llama-3-8B(S)      & 0.9062\colorbox{red!25}{(-0.005)}     & 0.9065\colorbox{red!25}{(-0.005)}     & 0.0000\colorbox{red!0}{(+0.000)}     \\
Qwen-7B(S)         & 0.6737\colorbox{green!25}{(+0.089)}   & 0.8226\colorbox{green!50}{(+0.164)}   & 0.1812\colorbox{green!25}{(-0.004)}   \\
Qwen-14B(S)        & 0.7887\colorbox{green!50}{(+0.131)}   & 0.8548\colorbox{green!50}{(+0.170)}   & 0.0775\colorbox{green!25}{(-0.003)}   \\
Vicuna-7B(S)       & 0.7925\colorbox{green!25}{(+0.083)}   & 0.7899\colorbox{green!25}{(+0.074)}   & 0.0000\colorbox{green!25}{(-0.005)}   \\
Vicuna-13B(S)      & 0.9075\colorbox{green!25}{(+0.071)}   & 0.9153\colorbox{green!25}{(+0.065)}   & 0.0088\colorbox{green!25}{(-0.005)}   \\
\hline
Llama-3-8B(F)  & 0.9175\colorbox{green!25}{(+0.006)} & 0.9164\colorbox{green!25}{(+0.003)} & 0.0000\colorbox{red!0}{(+0.000)} \\
Qwen-7B(F)  & \textbf{0.9713}\colorbox{green!75}{(+0.386)} & 0.9713\colorbox{green!75}{(+0.313)} & 0.0000\colorbox{green!50}{(-0.185)} \\
\bottomrule
\multicolumn{4}{l}{\footnotesize S: with few shot strategy; F: with fine-tuned strategy} \\
\end{tabular}
\end{table}

Table \ref{tab:e_commerce} presents results for the e-commerce product text classification dataset. The GRU model shows the best performance among all models except for fine-tuned LLMs with an accuracy of $0.9387$ and an F1 score of $0.9383$, making it the leading model in these categories before considering fine-tuning. This illustrates the capability of GRU to handle sequence and context effectively, which is crucial for product text classification. Like Table \ref{tab: twitters}, traditional algorithms exhibit much lower accuracy and F1 scores than NN-based models. Among LLMs, GPT-based models also show impressive results before fine-tuning, with GPT-3.5 achieving slightly higher metrics than GPT-4. Applying fine-tuning techniques to LLMs such as Qwen-7B can result in superior accuracy of $0.9713$ and F1 scores of $0.9713$, making these models particularly effective for specialized tasks such as e-commerce product text classification.

\begin{table}[h]
\caption{\small{ECONOMIC TEXTS Sentiment Classification Results}}
\centering
\label{tab: economic_texts}
\renewcommand{\arraystretch}{1.4}
\begin{tabular}{l@{\hspace{2mm}}c@{\hspace{2mm}}c@{\hspace{2mm}}c@{\hspace{0mm}}}
\hline
Model & ACC($\uparrow$) & F1($\uparrow$) & U/E($\downarrow$) \\ \hline
MNB  & 0.2600 & 0.2570 & - \\
LR   & 0.5962 & 0.3055 & - \\
RF   & 0.6375 & 0.4048 & - \\
DT   & 0.4813 & 0.3805 & - \\
KNN  & 0.5325 & 0.3528 & - \\
\hline
GRU & 0.6837 & 0.5494 & - \\
LSTM & 0.6950 & 0.5967 & - \\
RNN & 0.6550 & 0.4298 & - \\
\hline
BART     & 0.4125 & 0.4152 & - \\
DeBERTa  & 0.4025 & 0.4119 & - \\
\hline
GPT-3.5              & 0.6175 & 0.6063 & 0.0000 \\
GPT-4                & 0.7638 & 0.7659 & 0.0000 \\
Gemini-pro           & 0.7488 & 0.7519 & 0.0013 \\
Llama-3-8B           & 0.7675 & 0.7710 & 0.0013 \\
Qwen-7B              & 0.7550 & 0.7585 & 0.0025 \\
Qwen-14B             & 0.7850 & 0.7860 & 0.0050 \\
Vicuna-7B            & 0.7425 & 0.7250 & 0.0000 \\
Vicuna-13B           & 0.6750 & 0.6735 & 0.0013 \\
\hline
Gemini-pro(S)      & 0.6925\colorbox{red!25}{(-0.056)}     & 0.7217\colorbox{red!25}{(-0.030)}     & 0.0400\colorbox{red!25}{(+0.039)}     \\
Llama-3-8B(S)      & 0.7550\colorbox{red!25}{(-0.012)}     & 0.7585\colorbox{red!25}{(-0.013)}     & 0.0013\colorbox{red!25}{(+0.000)}     \\
Qwen-7B(S)         & 0.6837\colorbox{red!25}{(-0.071)}     & 0.6900\colorbox{red!25}{(-0.069)}     & 0.0288\colorbox{red!25}{(+0.026)}     \\
Qwen-14B(S)        & 0.7738\colorbox{red!25}{(-0.011)}     & 0.7748\colorbox{red!25}{(-0.011)}     & 0.0063\colorbox{red!25}{(+0.001)}     \\
Vicuna-7B(S)       & 0.7738\colorbox{green!25}{(+0.031)}   & 0.7607\colorbox{green!25}{(+0.036)}   & 0.0000\colorbox{red!0}{(+0.000)}     \\
Vicuna-13B(S)      & 0.7575\colorbox{green!25}{(+0.082)}   & 0.7616\colorbox{green!25}{(+0.088)}   & 0.0013\colorbox{red!0}{(+0.000)}     \\
\hline
Llama-3-8B & 0.7913\colorbox{green!25}{(+0.024)} & 0.7796\colorbox{green!25}{(+0.009)} & 0.0000\colorbox{green!25}{(-0.001)} \\
Qwen-7B(F) & \textbf{0.8400}\colorbox{green!25}{(+0.085)} & \textbf{0.8302}\colorbox{green!25}{(+0.074)} & 0.0000\colorbox{green!25}{(-0.003)} \\

\bottomrule
\multicolumn{4}{l}{\footnotesize S: with few shot strategy; F: with fine-tuned strategy} \\
\end{tabular}
\end{table}

Table \ref{tab: economic_texts} presents results for the economic texts sentiment classification dataset. The models show a broad performance spectrum, with the best results observed in fine-tuned LLMs. Traditional models continue to exhibit relatively low accuracy and F1 scores. RF performs somewhat better within this group but remains significantly lower than advanced models with a $0.6375$ accuracy and a $0.4048$ F1 score. NN-based models perform adequately, with GRU notably achieving an accuracy of $0.6837$ and an F1 score of $0.5494$. However, their performance is outstripped by more sophisticated models. LLM models, like GPT-4 and Gemini-pro, show significant improvements over traditional models, with GPT-4 reaching an accuracy of $0.7638$ and an F1 score of $0.7659$, indicating robust capabilities in processing complex economic texts. The fine-tuned models Llama-3-8B(F) and Qwen-7B(F) exhibit exceptional performance, with Qwen-7B(F) standing out for its remarkable accuracy and F1 score improvements. It is the only model that surpasses 80\% accuracy and F1 score.

\begin{table}[h]
\caption{\small{SMS SPAM COLLECTION Classification Results}}
\centering
\label{tab: sms_spam}
\renewcommand{\arraystretch}{1.4}
\begin{tabular}{l@{\hspace{2mm}}c@{\hspace{2mm}}c@{\hspace{2mm}}c@{\hspace{0mm}}}
\hline
Model & ACC($\uparrow$) & F1($\uparrow$) & U/E($\downarrow$) 
\\ \hline
MNB  & 0.7488 & 0.6376 & - \\
LR   & 0.8575 & 0.5419 & - \\
RF   & 0.8962 & 0.7196 & - \\
DT   & 0.8287 & 0.6559 & - \\
KNN  & 0.8237 & 0.6241 & - \\
\hline
GRU  & 0.9675 & 0.9257 & - \\
LSTM & 0.9675 & 0.9237 & - \\
RNN  & 0.9725 & 0.9366 & - \\
\hline
BART     & 0.7137 & 0.4943 & - \\
DeBERTa  & 0.7025 & 0.5630 & - \\
\hline
GPT-3.5                 & 0.4988 & 0.5601 & 0.0000 \\
GPT-4                   & 0.9463 & 0.9495 & 0.0000 \\
Gemini-pro              & 0.6500 & 0.7395 & 0.0575 \\
Llama-3-8B              & 0.3937 & 0.4426 & 0.0025 \\
Qwen-7B                 & 0.7050 & 0.7527 & 0.0013 \\
Qwen-14B                & 0.9137 & 0.9208 & 0.0000 \\
Vicuna-7B               & 0.2762 & 0.2847 & 0.0000 \\
Vicuna-13B              & 0.4550 & 0.5149 & 0.0000 \\
\hline
Gemini-pro(S)      & 0.8163\colorbox{green!50}{(+0.166)}   & 0.8759\colorbox{green!50}{(+0.136)}   & 0.0488\colorbox{green!25}{(-0.009)}   \\
Llama-3-8B(S)      & 0.5825\colorbox{green!50}{(+0.189)}   & 0.6482\colorbox{green!50}{(+0.206)}   & 0.0088\colorbox{red!25}{(+0.006)}     \\
Qwen-7B(S)         & 0.7525\colorbox{green!25}{(+0.047)}   & 0.8124\colorbox{green!25}{(+0.060)}   & 0.0362\colorbox{red!25}{(+0.035)}     \\
Qwen-14B(S)        & 0.8525\colorbox{red!25}{(-0.061)}     & 0.8730\colorbox{red!25}{(-0.048)}     & 0.0025\colorbox{red!25}{(+0.003)}     \\
Vicuna-7B(S)       & 0.5675\colorbox{green!75}{(+0.291)}   & 0.6310\colorbox{green!75}{(+0.346)}   & 0.0013\colorbox{red!25}{(+0.001)}     \\
Vicuna-13B(S)      & 0.6412\colorbox{green!50}{(+0.186)}   & 0.6976\colorbox{green!50}{(+0.183)}   & 0.0000\colorbox{red!0}{(+0.000)}     \\
\hline
Llama-3-8B(F) & 0.9825\colorbox{green!100}{(+0.589)} & 0.9826\colorbox{green!100}{(+0.540)} & 0.0000\colorbox{green!75}{(-0.003)} \\
Qwen-7B(F)  & \textbf{0.9938}\colorbox{green!75}{(+0.289)} & \textbf{0.9937}\colorbox{green!50}{(+0.241)} & 0.0000\colorbox{red!0}{(+0.000)} \\
\bottomrule
\multicolumn{4}{l}{\footnotesize S: with few shot strategy; F: with fine-tuned strategy} \\
\end{tabular}
\end{table}

Table \ref{tab: sms_spam} details the SMS spam collection classification results, showcasing a notable disparity in model effectiveness, with fine-tuned LLMs and NN-based models outperforming others by a wide margin. Once again, traditional models underperform compared to NN and some LLM models, with RF leading the traditional pack but not nearly matching the performance of advanced models. NN-based models show exceptionally high performance, with RNN achieving the best results with an accuracy of $0.9725$ and an F1 score of $0.9366$. For LLMs, while some models demonstrate their high abilities in detecting spam SMS with accuracy high to more than 90\%, like GPT-4 and Qwen-14B, some models failed in this task with accuracy lower than 0.5, like GPT-3.5, Llama-3-8B, and Vicuna families which are far worse than traditional ML methods or NN models. Notably, fine-tuning dramatically enhances the performance of models like Llama-3-8B(F) and Qwen-7B(F), which achieved the highest scores in both accuracy and F1 score with the values of $0.9938$ and $0.9927$, with the latter reaching near-perfect accuracy and F1 scores, highlighting the transformative power of model adaptation.

%% file: sec/discussion.tex
\section{Discussion}




\subsection{Prompting strategy}

The effectiveness of the few-shot strategy has been previously established; however, our investigation reveals that its influence is not uniform across different models and datasets.

In the context of Table \ref{tab: twitters}, five out of six models showed only marginal performance changes when employing this strategy. However, Qwen-7B(S) significantly underperformed with accuracy and F1 scores dropping by over 10\%. This trend was not mirrored in Table \ref{tab:e_commerce}, where four models marginally improved accuracy. Contrarily, Llama-3-8B(S) experienced a slight decrease, whereas Qwen-14B(S) notably excelled with an increase exceeding 13\%. Table \ref{tab: economic_texts} mostly saw marginal decreases in four out of six models, with only two showing minor improvements. These mixed results highlight that the impact of few-shot learning is highly model and dataset-dependent.

Table \ref{tab: sms_spam}, a different pattern emerged: while Qwen-7B(S) and Qwen-14B(S) underwent marginal changes in accuracy (4\% increase and 6\% decrease respectively), the other four models achieved significant improvements, Vicuna-7B(S), notably surged by over 25\%. As for U/E metrics across datasets, there were minor variations except for specific trends within each dataset; COVID-19-related tweets fluctuated both ways, e-commercial product texts predominantly decreased or remained unchanged, while Spam SMS and economic texts mostly saw increases. These observations underscore that while few-shot strategies can be potent tools for model enhancement, their application requires careful consideration of the interplay between model architectures and dataset nuances to harness their potential fully.

\subsection{Fine-tuning strategy}

Our research involved fine-tuning two LLMs across four datasets as presented in Table \ref{tab: twitters}, \ref{tab:e_commerce}, \ref{tab: economic_texts}, and \ref{tab: sms_spam}, with the results indicating a significant enhancement in text classification performance. Notably, the Llama-3-8B(F) model did not show an improvement in the COVID-19-related tweets sentiment dataset, as presented in Table V. However, this model's accuracy increased dramatically from 0.3937 to 0.9825 in spam SMS detection, transitioning from one of the least effective to one of the most proficient models, second only to Qwen-7B(F).

The Qwen-7B(F) model exhibited remarkable improvements across all datasets post-fine-tuning, with accuracy improved ranging from 0.085 to 0.386, thereby establishing it as a state-of-the-art model for these tasks. These findings highlight the potential of fine-tuning as a pivotal strategy for optimizing LLMs' performance on specific text classification tasks.

More importantly, after fine-tuning, the U/E value cross models and datasets are down to 0, except for the Llama-3-8B(F) in tweet classification. This improvement in the standardized output makes the result consistent and makes the system more reliable. 

Our results strongly advocate incorporating fine-tuning into LLM deployment workflows to unlock their full potential in specialized text classification scenarios. 

\subsection{Limitations}

While LLMs demonstrated impressive proficiency in text classification, our experiments also uncovered a range of limitations when leverage LLMs as text classifiers.

\begin{itemize}
    \item \textbf{Inconsistent Output Formats}: LLMs often produce inconsistent output formats, which can disrupt the integration into systems requiring standardized results (e.g., JSON format). This inconsistency challenges downstream applications that depend on structured data.

    \item \textbf{Content Classification Restrictions}: Some LLMs may refuse to classify certain types of content due to sensitivity or processing limitations, restricting their application scope in diverse or nuanced scenarios.

    \item \textbf{Proprietary Model Constraints}: Closed-source LLMs can limit scalability due to API rate limits and potentially prohibitive costs associated with high-volume usage, affecting real-time performance and accessibility.

    \item \textbf{Hardware Demands}: Utilizing LLMs, particularly larger models, requires significant CPU and GPU resources. This can hinder scalability and deployment in environments with limited access to high-performance computing.

    \item \textbf{Time-intensive Processing}: LLMs typically have longer inference times, impacting real-time or high-throughput applications. This trade-off between accuracy and efficiency is crucial for practitioners to consider.
\end{itemize}

%% file: main.bbl
\begin{thebibliography}{10}
\providecommand{\url}[1]{#1}
\csname url@samestyle\endcsname
\providecommand{\newblock}{\relax}
\providecommand{\bibinfo}[2]{#2}
\providecommand{\BIBentrySTDinterwordspacing}{\spaceskip=0pt\relax}
\providecommand{\BIBentryALTinterwordstretchfactor}{4}
\providecommand{\BIBentryALTinterwordspacing}{\spaceskip=\fontdimen2\font plus
\BIBentryALTinterwordstretchfactor\fontdimen3\font minus \fontdimen4\font\relax}
\providecommand{\BIBforeignlanguage}[2]{{%
\expandafter\ifx\csname l@#1\endcsname\relax
\typeout{** WARNING: IEEEtran.bst: No hyphenation pattern has been}%
\typeout{** loaded for the language `#1'. Using the pattern for}%
\typeout{** the default language instead.}%
\else
\language=\csname l@#1\endcsname
\fi
#2}}
\providecommand{\BIBdecl}{\relax}
\BIBdecl

\bibitem{liu2022sentiment}
B.~Liu, \emph{Sentiment analysis and opinion mining}.\hskip 1em plus 0.5em minus 0.4em\relax Springer Nature, 2022.

\bibitem{chen2020dirichlet}
J.~Chen, Z.~Gong, and W.~Liu, ``A dirichlet process biterm-based mixture model for short text stream clustering,'' \emph{Applied Intelligence}, vol.~50, pp. 1609--1619, 2020.

\bibitem{minaee2021deep}
S.~Minaee, N.~Kalchbrenner, E.~Cambria, N.~Nikzad, M.~Chenaghlu, and J.~Gao, ``Deep learning--based text classification: a comprehensive review,'' \emph{ACM computing surveys (CSUR)}, vol.~54, no.~3, pp. 1--40, 2021.

\bibitem{sarker2021machine}
I.~Sarker, ``Machine learning: algorithms, real-world applications and research directions. sn comput sci 2: 160,'' 2021.

\bibitem{wang2019survey}
W.~Wang, V.~W. Zheng, H.~Yu, and C.~Miao, ``A survey of zero-shot learning: Settings, methods, and applications,'' \emph{ACM Trans. on Intelligent Systems and Technology (TIST)}, vol.~10, no.~2, pp. 1--37, 2019.

\bibitem{chowdhery2023palm}
A.~Chowdhery, S.~Narang, J.~Devlin, M.~Bosma, G.~Mishra, A.~Roberts, P.~Barham, H.~W. Chung, C.~Sutton, S.~Gehrmann \emph{et~al.}, ``Palm: Scaling language modeling with pathways,'' \emph{Journal of Machine Learning Research}, vol.~24, no. 240, pp. 1--113, 2023.

\bibitem{touvron2023llama}
H.~Touvron, T.~Lavril, G.~Izacard, X.~Martinet, M.-A. Lachaux, T.~Lacroix, B.~Rozi{\`e}re, N.~Goyal, E.~Hambro, F.~Azhar \emph{et~al.}, ``Llama: Open and efficient foundation language models,'' \emph{arXiv preprint arXiv:2302.13971}, 2023.

\bibitem{radford2018improving}
A.~Radford, K.~Narasimhan, T.~Salimans, I.~Sutskever \emph{et~al.}, ``Improving language understanding by generative pre-training,'' 2018.

\bibitem{bețianu2024dallmi}
M.~Bețianu, A.~M{\u{a}}lan, M.~Aldinucci, R.~Birke, and L.~Chen, ``Dallmi: Domain adaption for llm-based multi-label classifier,'' in \emph{Pacific-Asia Conf. on Knwl. Discovery \& Data Mining}, 2024, pp. 277--289.

\bibitem{yan2023autocast}
Q.~Yan, R.~Seraj, J.~He, L.~Meng, and T.~Sylvain, ``Autocast++: Enhancing world event prediction with zero-shot ranking-based context retrieval,'' \emph{arXiv preprint arXiv:2310.01880}, 2023.

\bibitem{Wu2024llm}
X.~Wu, X.~Zhu, E.~Baralis, R.~Lu, V.~Kumar, L.~Rutkowski, and J.~Tang, ``On computing paradigms - where will large language models be going,'' in \emph{2024 IEEE Intl. Conference on Data Mining}, 2023, pp. 1577--1582.

\bibitem{quinlan2014c4}
J.~R. Quinlan, \emph{C4. 5: programs for machine learning}.\hskip 1em plus 0.5em minus 0.4em\relax Elsevier, 2014.

\bibitem{xu2018bayesian}
S.~Xu, ``Bayesian na{\"\i}ve bayes classifiers to text classification,'' \emph{Journal of Information Science}, vol.~44, no.~1, pp. 48--59, 2018.

\bibitem{rabiner1989tutorial}
L.~R. Rabiner, ``A tutorial on hidden markov models and selected applications in speech recognition,'' \emph{Proceedings of the IEEE}, vol.~77, no.~2, pp. 257--286, 1989.

\bibitem{joachims2002learning}
T.~Joachims, \emph{Learning to classify text using support vector machines}.\hskip 1em plus 0.5em minus 0.4em\relax Springer Science \& Business Media, 2002, vol. 668.

\bibitem{guo2003knn}
G.~Guo, H.~Wang, D.~Bell, Y.~Bi, and K.~Greer, ``Knn model-based approach in classification,'' in \emph{On The Move to Meaningful Internet Systems 2003: CoopIS, DOA, and ODBASE: OTM Confederated International Conferences, CoopIS, DOA, and ODBASE 2003, Catania, Sicily, Italy, November 3-7, 2003. Proceedings}.\hskip 1em plus 0.5em minus 0.4em\relax Springer, 2003, pp. 986--996.

\bibitem{genkin2007large}
A.~Genkin, D.~D. Lewis, and D.~Madigan, ``Large-scale bayesian logistic regression for text categorization,'' \emph{technometrics}, vol.~49, no.~3, pp. 291--304, 2007.

\bibitem{kim2014convolutional}
Y.~Kim, ``Convolutional neural networks for sentence classification,'' \emph{arXiv preprint arXiv:1408.5882}, 2014.

\bibitem{sutskever2014sequence}
I.~Sutskever, O.~Vinyals, and Q.~V. Le, ``Sequence to sequence learning with neural networks,'' \emph{Advances in neural information processing systems}, vol.~27, 2014.

\bibitem{chung2014empirical}
J.~Chung, C.~Gulcehre, K.~Cho, and Y.~Bengio, ``Empirical evaluation of gated recurrent neural networks on sequence modeling,'' \emph{arXiv preprint arXiv:1412.3555}, 2014.

\bibitem{zhou2024clcrnet}
W.~Zhou, L.~Zheng, X.~Li, Z.~Yang, and J.~Yi, ``Clcrnet: An optical character recognition network for cigarette laser code,'' \emph{IEEE Transactions on Instrumentation and Measurement}, 2024.

\bibitem{devlin2018bert}
J.~Devlin, M.-W. Chang, K.~Lee, and K.~Toutanova, ``Bert: Pre-training of deep bidirectional transformers for language understanding,'' \emph{arXiv preprint arXiv:1810.04805}, 2018.

\bibitem{raza2024nbias}
S.~Raza, M.~Garg, D.~J. Reji, S.~R. Bashir, and C.~Ding, ``Nbias: A natural language processing framework for bias identification in text,'' \emph{Expert Systems with Applications}, vol. 237, p. 121542, 2024.

\bibitem{zhou2024cdgan}
N.~Zhou, N.~Yao, N.~Hu, J.~Zhao, and Y.~Zhang, ``Cdgan-bert: Adversarial constraint and diversity discriminator for semi-supervised text classification,'' \emph{Knowledge-Based Systems}, vol. 284, p. 111291, 2024.

\bibitem{jamshidi2024effective}
S.~Jamshidi, M.~Mohammadi, S.~Bagheri, H.~E. Najafabadi, A.~Rezvanian, M.~Gheisari, M.~Ghaderzadeh, A.~S. Shahabi, and Z.~Wu, ``Effective text classification using bert, mtm lstm, and dt,'' \emph{Data \& Knowledge Engineering}, p. 102306, 2024.

\bibitem{brown2020language}
T.~Brown, B.~Mann, N.~Ryder, M.~Subbiah, J.~D. Kaplan, P.~Dhariwal, A.~Neelakantan, P.~Shyam, G.~Sastry, A.~Askell \emph{et~al.}, ``Language models are few-shot learners,'' \emph{Advances in neural information processing systems}, vol.~33, pp. 1877--1901, 2020.

\bibitem{raffel2020exploring}
C.~Raffel, N.~Shazeer, A.~Roberts, K.~Lee, S.~Narang, M.~Matena, Y.~Zhou, W.~Li, and P.~J. Liu, ``Exploring the limits of transfer learning with a unified text-to-text transformer,'' \emph{Journal of machine learning research}, vol.~21, no. 140, pp. 1--67, 2020.

\bibitem{peng2023rwkv}
B.~Peng, E.~Alcaide, Q.~Anthony, A.~Albalak, S.~Arcadinho, H.~Cao, X.~Cheng, M.~Chung, M.~Grella, K.~K. GV \emph{et~al.}, ``Rwkv: Reinventing rnns for the transformer era,'' \emph{arXiv preprint arXiv:2305.13048}, 2023.

\bibitem{gu2023mamba}
A.~Gu and T.~Dao, ``Mamba: Linear-time sequence modeling with selective state spaces,'' \emph{arXiv preprint arXiv:2312.00752}, 2023.

\bibitem{team2023gemini}
G.~Team, R.~Anil, S.~Borgeaud, Y.~Wu, J.-B. Alayrac, J.~Yu, R.~Soricut, J.~Schalkwyk, A.~M. Dai, A.~Hauth \emph{et~al.}, ``Gemini: a family of highly capable multimodal models,'' \emph{arXiv preprint arXiv:2312.11805}, 2023.

\bibitem{anthropic2024claude}
A.~Anthropic, ``The claude 3 model family: Opus, sonnet, haiku,'' \emph{Claude-3 Model Card}, 2024.

\bibitem{chae2023large}
Y.~Chae and T.~Davidson, ``Large language models for text classification: From zero-shot learning to fine-tuning,'' \emph{Open Science Foundation}, 2023.

\bibitem{loukas2023making}
L.~Loukas, I.~Stogiannidis, O.~Diamantopoulos, P.~Malakasiotis, and S.~Vassos, ``Making llms worth every penny: Resource-limited text classification in banking,'' in \emph{Proceedings of the Fourth ACM International Conference on AI in Finance}, 2023, pp. 392--400.

\bibitem{10386911}
F.~Wei, R.~Keeling, N.~Huber-Fliflet, J.~Zhang, A.~Dabrowski, J.~Yang, Q.~Mao, and H.~Qin, ``Empirical study of llm fine-tuning for text classification in legal document review,'' in \emph{2023 IEEE International Conference on Big Data (BigData)}, 2023, pp. 2786--2792.

\bibitem{10386772}
N.~A. Cloutier and N.~Japkowicz, ``Fine-tuned generative llm oversampling can improve performance over traditional techniques on multiclass imbalanced text classification,'' in \emph{2023 IEEE International Conference on Big Data (BigData)}, 2023, pp. 5181--5186.

\bibitem{wang2023large}
Z.~Wang, Y.~Pang, and Y.~Lin, ``Large language models are zero-shot text classifiers,'' \emph{arXiv preprint arXiv:2312.01044}, 2023.

\bibitem{gabrielpreda_2020}
\BIBentryALTinterwordspacing
G.~Preda, ``Covid19 tweets,'' 2020. [Online]. Available: \url{https://www.kaggle.com/dsv/1451513}
\BIBentrySTDinterwordspacing

\bibitem{malo2014good}
P.~Malo, A.~Sinha, P.~Korhonen, J.~Wallenius, and P.~Takala, ``Good debt or bad debt: Detecting semantic orientations in economic texts,'' \emph{J. of the Association for Info. Sci. \& Tech.}, vol.~65, no.~4, pp. 782--796, 2014.

\bibitem{gautam_2019_335582}
\BIBentryALTinterwordspacing
Gautam, ``E commerce text dataset,'' 2019. [Online]. Available: \url{https://doi.org/10.5281/zenodo.3355823}
\BIBentrySTDinterwordspacing

\bibitem{misc_sms_spam_collection_228}
T.~Almeida and J.~Hidalgo, ``{SMS Spam Collection},'' UCI Machine Learning Repository, 2012, {DOI}: https://doi.org/10.24432/C5CC84.

\end{thebibliography}
